\documentclass[conference]{IEEEtran}
\IEEEoverridecommandlockouts
\usepackage{cite}
\usepackage{amsmath,amssymb,amsfonts}
\usepackage{algorithmic}
\usepackage{graphicx}
\usepackage{textcomp}
\usepackage{xcolor}
\usepackage{array}
\usepackage{tikz}
\usepackage{adjustbox}
\usepackage{fontawesome5}
\usepackage[edges]{forest}
\tikzset{%
    parent/.style =          {align=center,text width=3.5cm,rounded corners=3pt},
    child/.style =           {align=center,text width=3.5cm,rounded corners=3pt},
    grandchild/.style =      {align=center,text width=3.5cm,rounded corners=3pt},
    greatgrandchild/.style = {align=center,text width=1.5cm,rounded corners=3pt},
    referenceblock/.style =  {align=center,text width=1.5cm,rounded corners=2pt}
}
\usepackage{float}
\usepackage{placeins}

\usepackage{comment}
\def\BibTeX{{\rm B\kern-.05em{\sc i\kern-.025em b}\kern-.08em
    T\kern-.1667em\lower.7ex\hbox{E}\kern-.125emX}}
\begin{document}

\title{A Survey of Large Language Model-Based Generative AI for Text-to-SQL: Benchmarks, Applications, Use Cases, and Challenges}

\author{
\IEEEauthorblockN{Aditi Singh\textsuperscript{1}, Akash Shetty\textsuperscript{1}, Abul Ehtesham\textsuperscript{2}, Saket Kumar\textsuperscript{3}, Tala Talaei Khoei\textsuperscript{4}}
\IEEEauthorblockA{
\textsuperscript{1}\textit{Department of Computer Science, Cleveland State University, USA} \\
\textsuperscript{2}\textit{The Davey Tree Expert Company, USA} \\
\textsuperscript{3}\textit{The Mathworks, USA} \\
\textsuperscript{4}\textit{Khoury College of Computer Science, 
Roux Institute at Northeastern University,  USA} \\
a.singh22@csuohio.edu, a.shetty13@vikes.csuohio.edu, abul.ehtesham@davey.com,  \\ 
saketk@mathworks.com, t.talaeikhoei@northeastern.edu
}
}

\maketitle

\begin{abstract}
Text-to-SQL systems facilitate smooth interaction with databases by translating natural language queries into Structured Query Language (SQL), bridging the gap between non-technical users and complex database management systems. This survey provides a comprehensive overview of the evolution of AI-driven text-to-SQL systems, highlighting their foundational components, advancements in large language model (LLM) architectures, and the critical role of datasets such as Spider, WikiSQL, and CoSQL in driving progress.
We examine the applications of text-to-SQL in domains like healthcare, education, and finance, emphasizing their transformative potential for improving data accessibility. Additionally, we analyze persistent challenges, including domain generalization, query optimization, support for multi-turn conversational interactions, and the limited availability of datasets tailored for NoSQL databases and dynamic real-world scenarios.
To address these challenges, we outline future research directions, such as extending text-to-SQL capabilities to support NoSQL databases, designing datasets for dynamic multi-turn interactions, and optimizing systems for real-world scalability and robustness. By surveying current advancements and identifying key gaps, this paper aims to guide the next generation of research and applications in LLM-based text-to-SQL systems.
\end{abstract}


\section{Introduction}
The task of translating natural language questions into Structured Query Language (SQL) statements, known as text-to-SQL, has garnered significant attention within the fields of natural language processing and database management. This capability democratizes data access and analysis, enabling users to interact with databases without requiring in-depth knowledge of query languages. The development of AI-driven text-to-SQL systems has been critical in achieving this goal.\cite{yu2018spider}

Early approaches to text-to-SQL relied heavily on rule-based systems and semantic parsing techniques. These methods, while foundational, often struggled with the diversity and complexity inherent in natural language queries. The advent of deep learning and neural network models marked a significant shift, introducing sequence-to-sequence architectures that improved the translation of natural language to SQL\cite{shi2024survey}.

The integration of large pre-trained language models (PLMs) and large language models has further advanced the field\cite{singh2023}, enhancing the understanding of natural language semantics and the generation of accurate SQL queries. Recent surveys have highlighted the impact of PLMs on text-to-SQL parsing, noting their ability to capture complex linguistic patterns and improve performance across  benchmarks\cite{hong2024next}.

Despite these advancements, challenges remain, particularly in handling complex and cross-domain queries. The development of large-scale, human-labeled datasets, such as Spider, has been instrumental in evaluating and advancing text-to-SQL systems. These datasets provide diverse and complex queries that test the robustness and adaptability of current models.\cite{yu2018spider}.

This survey aims to provide a comprehensive overview of the evolution of text-to-SQL systems, emphasizing the integration of artificial intelligence methodologies. We explore foundational concepts, current benchmarks, datasets, and models, offering insights into the advancements and challenges in the field. By examining the trajectory of text-to-SQL research, we aim to highlight the progress made and identify areas for future exploration.

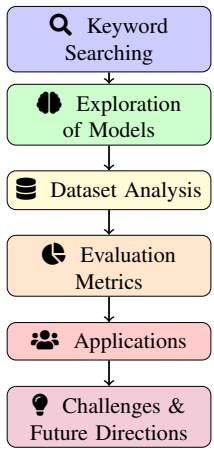
\begin{figure}[!ht]
\centering
\begin{adjustbox}{width=0.15\textwidth}
\begin{tikzpicture}[node distance=1.2cm and 0.6cm, every node/.style={align=center}]
    \node (keywords) [rectangle, draw, rounded corners, fill=blue!20, text width=3cm, minimum height=0.4cm] {\faSearch ~ Keyword Searching};
    \node (models) [rectangle, draw, rounded corners, fill=green!20, text width=3cm, below of=keywords] {\faBrain ~ Exploration of Models};
    \node (datasets) [rectangle, draw, rounded corners, fill=yellow!20, text width=3cm, below of=models] {\faDatabase ~ Dataset Analysis};
    \node (metrics) [rectangle, draw, rounded corners, fill=orange!20, text width=3cm, below of=datasets] {\faChartPie ~ Evaluation Metrics};
    \node (applications) [rectangle, draw, rounded corners, fill=red!20, text width=3cm, below of=metrics] {\faUsers ~ Applications};
    \node (challenges) [rectangle, draw, rounded corners, fill=purple!20, text width=3cm, below of=applications] {\faLightbulb ~ Challenges \& Future Directions};

    \draw[->, thick] (keywords) -- (models);
    \draw[->, thick] (models) -- (datasets);
    \draw[->, thick] (datasets) -- (metrics);
    \draw[->, thick] (metrics) -- (applications);
    \draw[->, thick] (applications) -- (challenges);
\end{tikzpicture}
\end{adjustbox}
\caption{Methodology for conducting the survey of Text-to-SQL systems.}
\label{fig:methodology}
\end{figure}

\section{Need for Text-to-SQL} 
Text-to-SQL systems provide a specialized solution for translating natural language queries into precise SQL statements, allowing users to interact directly with databases without requiring expertise in SQL syntax. While general-purpose AI models like ChatGPT can assist in generating SQL queries, they often lack the domain-specific optimizations and accuracy that dedicated text-to-SQL systems are designed to offer. These systems are tailored to manage complex database schemas and ensure the generation of syntactically correct and efficient SQL queries, significantly enhancing data retrieval and analysis processes.
By focusing exclusively on the task of converting natural language to SQL, text-to-SQL systems deliver more reliable and contextually appropriate results. This makes them invaluable in scenarios where precise data manipulation is critical, such as in healthcare, finance, and business intelligence. Furthermore, the development of such systems incorporates advancements in natural language understanding, database schema modeling, and semantic parsing, contributing to their robustness and usability across diverse application domains.

\section{Foundations of Text-to-SQL}

Text-to-SQL systems are designed to translate natural language queries into Structured Query Language (SQL) statements, enabling users to interact with databases without requiring expertise in SQL syntax. The foundational components of these systems as shown in Figure \ref{fig:Text-to-SQL Process Overview} include:

\begin{figure}[!h] 
    \centering
    \includegraphics[width=0.45\textwidth]{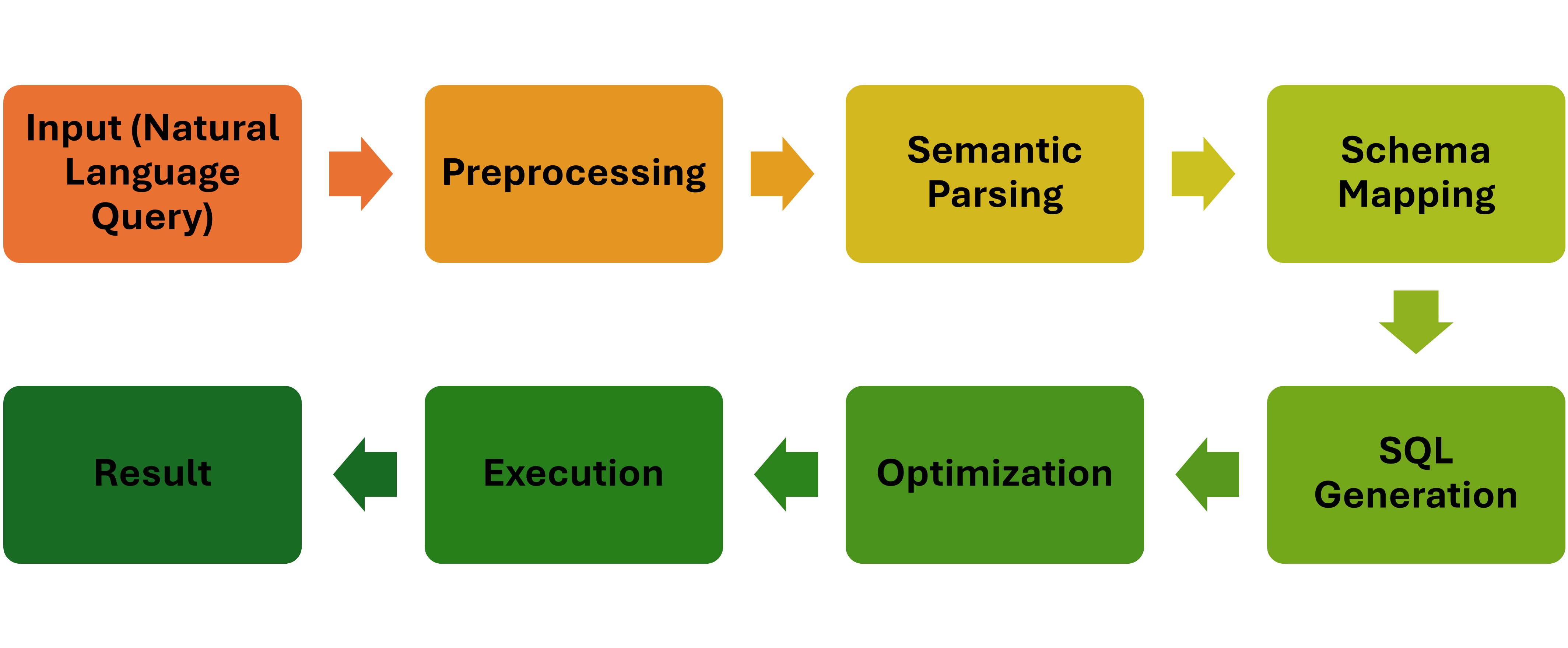} 
    \caption{Text-to-SQL Process Overview}
    \label{fig:Text-to-SQL Process Overview}
\end{figure}

\begin{figure*}[!ht] 
    \centering
    \includegraphics[width=0.6\textwidth]{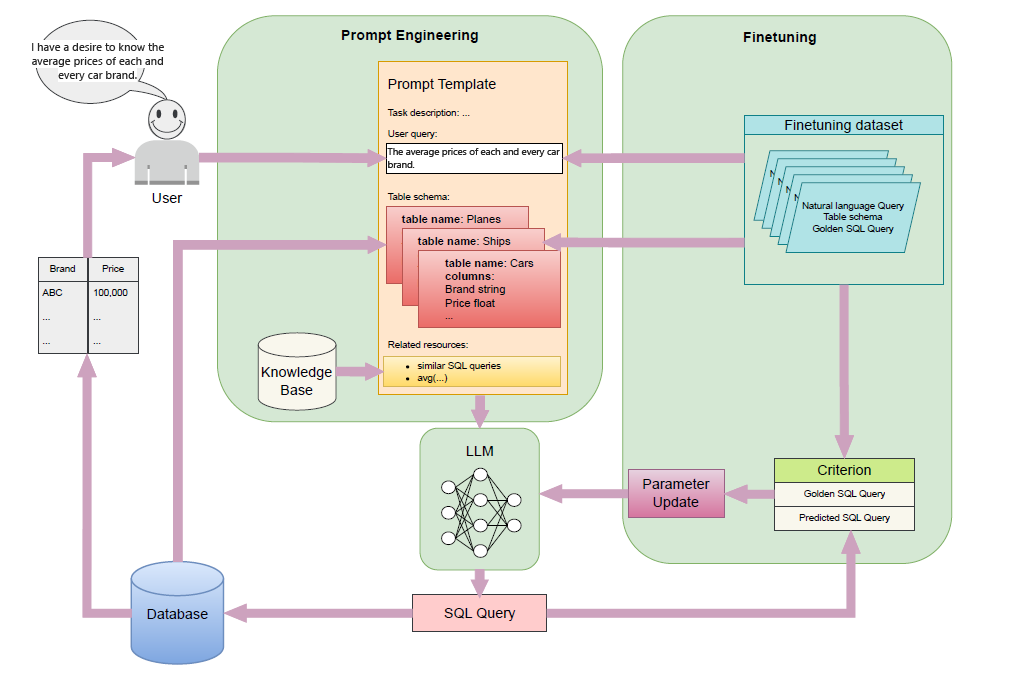} 
    \caption{LLM Framework for Text-to-SQL \cite{shi2024surveyemployinglargelanguage}}
    \label{fig:Framework of employing LLMs in Text-to-SQL}
\end{figure*}

\begin{enumerate} \item \textbf{Natural Language Understanding (NLU):} This involves parsing and interpreting the user's query to comprehend its intent and semantics. Techniques such as tokenization, part-of-speech tagging, and syntactic parsing are employed to analyze the structure and meaning of the input.
\item \textbf{Schema Linking:} This process connects elements of the natural language query to the corresponding components in the database schema, such as tables and columns. Effective schema linking is crucial for accurately mapping user intents to database structures.

\item \textbf{Semantic Parsing:} This step involves converting the natural language query into an intermediate logical form that represents its meaning. Semantic parsing serves as a bridge between the user's intent and the formal SQL query.

\item \textbf{SQL Generation:} The final component translates the intermediate logical form into a syntactically correct and executable SQL statement. This requires understanding SQL syntax and ensuring that the generated query aligns with the database schema.

\end{enumerate}

Advancements in artificial intelligence, particularly in deep learning and natural language processing, have significantly enhanced the performance of text-to-SQL systems. For instance, the integration of large pre-trained language models has improved the systems' ability to understand complex queries and generate accurate SQL statements \cite{hong2024next}.

Despite these advancements, challenges remain, especially in handling complex and cross-domain queries. Ongoing research focuses on improving the robustness and adaptability of text-to-SQL systems to address these challenges \cite{shi2024survey}.

\section{Current Benchmarks, Models, and Datasets}
Evaluating text-to-SQL systems necessitates robust benchmarks and datasets. Notable among these are:

\subsection{Benchmarks Datasets}
\begin{itemize}
    \item \textbf{Spider}: A large-scale, complex, and cross-domain text-to-SQL dataset designed to evaluate the generalization capabilities of models across different databases and query structures \cite{yu2018spider}.
    \item \textbf{Spider 2.0}: An advanced evaluation framework featuring 632 real-world text-to-SQL workflow problems from enterprise databases. These databases, often hosted on platforms like BigQuery and Snowflake, include over 1,000 columns. Spider 2.0 challenges models with complex tasks requiring interaction with SQL workflows, reasoning over extensive contexts, and generating multi-query SQL operations exceeding 100 lines, making it essential for assessing language models in enterprise scenarios \cite{lei2024spider}. 
    \item \textbf{WikiSQL}: Comprising over 80,000 natural language questions and corresponding SQL queries, this dataset is derived from Wikipedia tables and focuses on simple SQL queries \cite{zhong2017seq2sql}.
    \item \textbf{BIRD (BIg Bench for LaRge-Scale Database Grounded Text-to-SQL Evaluation)}: A comprehensive dataset containing 12,751 question-SQL pairs across 95 databases, totaling 33.4 GB. It spans over 37 professional domains, including blockchain, hockey, healthcare, and education, emphasizing challenges such as handling extensive database contents and integrating external knowledge \cite{li2023bird}.
    \item \textbf{CSpider}: A Chinese large-scale, complex, cross-domain text-to-SQL dataset, translated from the original Spider dataset. It comprises 10,181 questions and 5,693 unique SQL queries across 200 databases, aiming to facilitate the development of natural language interfaces for Chinese databases\cite{min2019pilotstudychinesesql}.
    \item \textbf{UNITE}: A unified benchmark composed of 18 publicly available text-to-SQL datasets, encompassing natural language questions from more than 12 domains, SQL queries from over 3,900 patterns, and 29,000 databases. It introduces approximately 120,000 additional examples and a threefold increase in SQL patterns compared to the Spider benchmark \cite{lan2023unite}.
    \item \textbf{CoSQL}: The CoSQL dataset is a dialogue-based benchmark designed for multi-turn text-to-SQL interactions. It comprises over 30,000 turns and more than 10,000 annotated SQL queries, collected from 3,000 dialogues across 200 complex databases spanning 138 domains. Unlike static text-to-SQL datasets, CoSQL emphasizes natural conversational interactions, simulating real-world scenarios where users refine, clarify, and expand their queries. This makes CoSQL a critical resource for advancing dialogue-based database interfaces. \cite{yu2019CoSQL}. 
\end{itemize}

\subsection{Models}
The progression of text-to-SQL models has been marked by several key developments (Table \ref{tab:models}): 

\begin{table*}[t]
\centering
\caption{Comparison of Text-to-SQL Models}
\renewcommand{\arraystretch}{1.5} 
\setlength{\tabcolsep}{4pt} 
\begin{tabular}{|p{4cm}|p{4cm}|p{6cm}|p{3cm}|}
\hline
\textbf{Model Name} & \textbf{Dataset} & \textbf{Training Method} & \textbf{Accuracy} \\ \hline
Seq2SQL & WikiSQL & Seq-to-Seq with Reinforcement Learning & 59.4\% \\ \hline
SQLNet & WikiSQL & Sketch-Based with Column Attention & 63.2\% \\ \hline
TypeSQL & WikiSQL & Type-Aware Neural Network & 82.6\% \\ \hline
IRNet & Spider & Graph Encoder + Intermediate Representation & 61.9\% \\ \hline
T5-3B & Spider, CoSQL & Fine-Tuned Transformer & 70.0\% \\ \hline
PICARD + T5-3B & CoSQL & Constrained Decoding for Dialogue-Based SQL Generation & High \\ \hline
RASAT+PICARD & CoSQL & Relation-Aware Self-Attention-augmented T5 with Incremental Parsing & 37.4\% IEX \\ \hline
MedT5SQL & MIMICSQL & BERT-based Encoder with LSTM Decoder for SQL Translation & High Accuracy in Medical Query Translation \\ \hline
EDU-T5 & Custom Educational Dataset & Fine-tuned T5 Model with Cross-Attention for SQL Query Generation & Optimized \\ \hline
RAT-SQL & WikiSQL, Spider & Relation-Aware Transformer & 69.7\% \\ \hline
SQLova & WikiSQL & BERT + Column Attention & 95\% \\ \hline
X-SQL & WikiSQL & BERT-style pre-training with context & 91.8\% \\ \hline
EHRSQL & EHRSQL Benchmark & Benchmark Model for EHRs & N/A \\ \hline
\end{tabular}
\label{tab:models}
\end{table*}
\begin{itemize}
    \item \textbf{Seq2SQL}: An early model that employs a sequence-to-sequence approach with reinforcement learning to generate SQL queries from natural language \cite{zhong2017seq2sql}.
    \item \textbf{SQLNet}: Introduces a sketch-based approach to predict the SQL query structure before filling in specific details, improving accuracy and efficiency \cite{xu2017sqlnet}.
    \item \textbf{TypeSQL}: Enhances SQLNet by incorporating type information, enabling the model to handle more complex queries involving different data types \cite{yu2018typesql}.
    \item \textbf{IRNet}: Utilizes a graph-based encoder to capture the relationships between database schema elements and natural language questions, leading to improved performance on complex queries \cite{GuoIRNet2019}.
    \item \textbf{T5-3B}: A transformer-based model fine-tuned on text-to-SQL tasks, demonstrating significant improvements in generating accurate SQL queries \cite{raffel2020exploring}.
    \item \textbf{MedT5SQL}: Tailored for healthcare, MedTS generates SQL queries for patient records using a BERT-based encoder and LSTM decoder trained on the MIMICSQL dataset. \cite{marshan2024medt5sql}. 
    \item \textbf{EDU-T5}: Optimized for educational data, EDU-T5 translates academic queries into SQL, using a T5-based model with cross-attention mechanisms. \cite{raffel2020exploring}. 
    \item \textbf{SQLova}: Built on WikiSQL, SQLova generates high-precision general-purpose SQL queries by combining a BERT-based encoder and column attention. \cite{hwang2019comprehensiveexplorationwikisqltableaware}. 
    \item \textbf{RAT-SQL}: Trained on WikiSQL and Spider, RAT-SQL uses a relation-aware transformer with schema encoding to manage complex multi-table queries. \cite{wang2021ratsqlrelationawareschemaencoding}. 
    \item \textbf{X-SQL}: Enhances schema representation by integrating contextual outputs from BERT-style models, achieving state-of-the-art performance on the WikiSQL dataset. \cite{he2019xsqlreinforceschemarepresentation}.
    \item \textbf{EHRSQL}: A benchmark designed for generating SQL queries from electronic health records, emphasizing domain-specific challenges and evaluation \cite{lee2023ehrsql}.
    \item \textbf{RASAT}: A relation-aware self-attention transformer model optimized for complex queries and integrated with dialogue-based datasets like CoSQL \cite{qi2022rasatintegratingrelationalstructures}.
    \item \textbf{PICARD}: Parsing incrementally for constrained auto-regressive decoding, PICARD improves the performance of language models like T5-3B on dialogue-based and multi-turn SQL generation tasks \cite{scholak2021picardparsingincrementallyconstrained}.
\end{itemize}

\subsection{Evaluation Metrics}
The performance of Text-to-SQL systems is assessed using several evaluation metrics that capture different aspects of query generation quality and system usability:
\begin{itemize}
    \item \textbf{Exact Set Match Accuracy:} Measures the proportion of SQL queries that match the ground truth exactly, ensuring structural correctness\cite{ascoli2024esmmoderninsightsperspective}.
    \item \textbf{Execution Accuracy:} Evaluates the system's ability to generate executable queries by checking if the generated query returns the correct results when executed on the database \cite{ascoli2024esmmoderninsightsperspective}.
    \item \textbf{Question Match Accuracy:} Assesses how well the generated query corresponds to the natural language question \cite{ascoli2024esmmoderninsightsperspective}. 
    \item \textbf{Interaction Match Accuracy:} Specifically for multi-turn datasets, measures the system's ability to maintain context and generate coherent queries across dialogue turns \cite{{ascoli2024esmmoderninsightsperspective}}.
\end{itemize}
These metrics provide insights into the accuracy, robustness, and practical usability of Text-to-SQL systems across different scenarios.

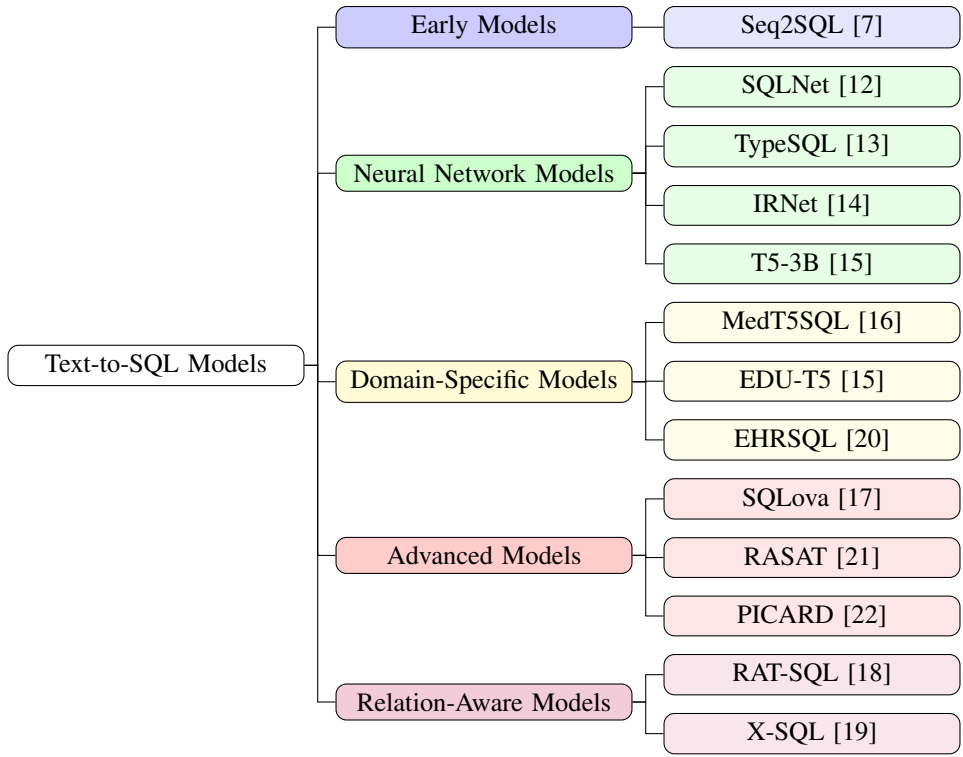
\begin{figure*}[htbp]
\centering
\begin{forest}
for tree={%
                forked edges,
                grow'=0,
                draw,
                rounded corners,
                node options={align=center,},
                text width=3.7cm,
            },
[Text-to-SQL Models
    [Early Models, fill=blue!20
        [Seq2SQL \cite{zhong2017seq2sql}, fill=blue!10]
    ]
    [Neural Network Models, fill=green!20
        [SQLNet \cite{xu2017sqlnet}, fill=green!10]
        [TypeSQL \cite{yu2018typesql}, fill=green!10]
        [IRNet \cite{GuoIRNet2019}, fill=green!10]
        [T5-3B \cite{raffel2020exploring}, fill=green!10]
    ]
    [Domain-Specific Models, fill=yellow!20
        [MedT5SQL \cite{marshan2024medt5sql}, fill=yellow!10]
        [EDU-T5 \cite{raffel2020exploring}, fill=yellow!10]
        [EHRSQL \cite{lee2023ehrsql}, fill=yellow!10]
    ]
    [Advanced Models, fill=red!20
        [SQLova \cite{hwang2019comprehensiveexplorationwikisqltableaware}, fill=red!10]
        [RASAT \cite{qi2022rasatintegratingrelationalstructures}, fill=red!10]
        [PICARD \cite{scholak2021picardparsingincrementallyconstrained}, fill=red!10]
    ]
    [Relation-Aware Models, fill=purple!20
        [RAT-SQL \cite{wang2021ratsqlrelationawareschemaencoding}, fill=purple!10]
        [X-SQL \cite{he2019xsqlreinforceschemarepresentation}, fill=purple!10]
    ]
]
\end{forest}
\caption{Hierarchical Tree of Text-to-SQL Models}
\label{fig:forest_models}
\end{figure*}

\section{Application and Use Cases}

Text-to-SQL systems serve as a pivotal tool across diverse industries, enabling natural language interaction with databases to facilitate data retrieval, analysis, and decision-making. These systems simplify complex data queries, empowering non-technical users to unlock valuable insights from structured data. Table \ref{tab:applications} summarizes the practical applications, challenges, and benefits of Text-to-SQL systems across key domains. Below, we detail specific applications in healthcare, education, finance, and business intelligence, showcasing the transformative potential of these systems in addressing industry-specific challenges.

\begin{table*}[!h]
\centering
\caption{Applications, Challenges, and Benefits of Text-to-SQL}
\renewcommand{\arraystretch}{1.3} 
\setlength{\tabcolsep}{4pt}
\begin{tabular}{|>{\centering\arraybackslash}m{3cm}|>{\centering\arraybackslash}m{6.5cm}|>{\centering\arraybackslash}m{6.5cm}|} 
\hline
\textbf{Applications} & \textbf{Challenges} & \textbf{Benefits of Text-to-SQL} \\ \hline

\textbf{Healthcare} &
\begin{itemize}
    \item Complex schemas involving multiple interconnected tables for patient, treatment, and diagnosis data.
    \item Domain-specific terminology and abbreviations requiring specialized understanding.
    \item Integration of external medical knowledge (e.g., guidelines, disease ontologies) for accurate responses.
\end{itemize} &
\begin{itemize}
    \item Simplifies access to patient records, aiding clinicians and healthcare staff.
    \item Improves decision-making through faster data retrieval for evidence-based practices.
    \item Automates administrative tasks, reducing workload and errors.
\end{itemize} \\ \hline

\textbf{Education} &
\begin{itemize}
    \item Diverse data formats across institutions, such as course catalogs and grade records.
    \item Ambiguity in queries due to varying terminologies among educators and students.
    \item Adapting to multiple educational levels and domains (e.g., K-12 vs. higher education).
\end{itemize} &
\begin{itemize}
    \item Facilitates analytics on academic performance, aiding in identifying strengths and weaknesses.
    \item Supports personalized learning by analyzing individual student progress.
    \item Scales efficiently to handle large datasets, such as nationwide assessments.
\end{itemize} \\ \hline

\textbf{Finance} &
\begin{itemize}
    \item High query complexity due to multi-faceted financial transactions.
    \item Ambiguity in fraud detection rules and terminology inconsistencies across organizations.
    \item Need for efficient query execution in real-time analytics scenarios.
\end{itemize} &
\begin{itemize}
    \item Enhances fraud detection by enabling effective querying of transaction data.
    \item Provides real-time insights for financial reporting and decision-making.
    \item Improves risk management by analyzing transactional and market data effectively.
\end{itemize} \\ \hline

\textbf{General Applications} &
\begin{itemize}
    \item Lack of generalization across diverse domains and transfer learning capabilities.
    \item Interpretability issues in generated queries.
    \item Debugging challenges when queries yield unexpected results or errors.
\end{itemize} &
\begin{itemize}
    \item Enables universal SQL query generation across multiple domains.
    \item Improves accessibility for non-technical users to interact with databases.
    \item Enhances accuracy and efficiency in data retrieval tasks.
\end{itemize} \\ \hline

\end{tabular}
\label{tab:applications}
\end{table*}

\subsection{Healthcare and Medical Records - Data Management}
\begin{itemize}
    \item \textbf{Clinical Decision Support}: Text-to-SQL systems assist healthcare professionals in querying patient records, retrieving relevant medical histories, or aggregating patient data for epidemiological studies.
    \item \textbf{Patient Record Management}: Allows healthcare providers to retrieve complex patient information, such as identifying all patients with certain conditions or recent test results, enabling faster access to critical data.
    \item \textbf{Medical Research}: Facilitates large-scale data retrieval from medical databases for research on disease patterns, drug efficacy, or population health studies.
\end{itemize}

\subsection{Educational Tool}
\begin{itemize}
    \item \textbf{Adaptive Learning Systems}: Text-to-SQL can be integrated into educational platforms to create adaptive learning tools that personalize content based on student data stored in databases.
    \item \textbf{Academic Analytics}: Enables querying of student performance data, helping in curriculum assessment, identifying at-risk students, and evaluating the impact of educational interventions.
    \item \textbf{Student Self-Service}: Allows students to query databases for information on course requirements, academic records, or library resources without needing technical expertise.
\end{itemize}

\subsection{Finance and Banking}
\begin{itemize}
    \item \textbf{Financial Reporting and Analytics}: Enables financial analysts and executives to interact with databases for generating real-time financial reports, trend analyses, and risk assessments.
    \item \textbf{Customer Service and Query Resolution}: Empowers customer service agents to query customer data efficiently, providing personalized and accurate information in real-time.
    \item \textbf{Fraud Detection and Prevention}: Allows for rapid query formulation to monitor transactions for patterns indicative of fraud, ensuring timely responses to suspicious activities.
\end{itemize}

\subsection{Business Intelligence and Analytics}
\begin{itemize}
    \item \textbf{Market Analysis and Trend Detection}: Text-to-SQL streamlines data retrieval for analyzing customer behavior, tracking sales performance, and identifying emerging market trends.
    \item \textbf{Inventory and Supply Chain Management}: Assists businesses in monitoring inventory levels, supply chain metrics, and order statuses to improve operational efficiency.
    \item \textbf{Employee Productivity and HR Analytics}: Enables querying workforce management data, allowing companies to monitor employee productivity, analyze turnover rates, or assess training needs.
    \item \textbf{LinkedIn’s SQL Bot - Data Democratization}: A multi-agent system built on LangChain and LangGraph, SQL Bot empowers LinkedIn employees to transform natural language questions into SQL queries, automating workflows, reducing dependencies on data teams, and accelerating decision-making across diverse business functions \cite{linkedin_sqlbot}.
\end{itemize}

\section{Challenges and Future Directions}

Despite significant advancements, several challenges remain in the development of text-to-SQL systems. Future research aims to address these challenges through the following directions.

\subsection{Industry-Specific Challenges and Solutions}
\textbf{Healthcare}: 
\begin{itemize}
    \item Handling highly sensitive patient data requires robust privacy-preserving techniques in text-to-SQL systems.
    \item Queries often require contextual medical knowledge not present in the database schema.
\end{itemize}

\textbf{Finance}:
\begin{itemize}
    \item The prevalence of NoSQL databases adds complexity to adapting text-to-SQL systems.
    \item Ambiguities in financial terminologies, such as risk-related queries, require domain-specific language models.
\end{itemize}

\textbf{Education}:
\begin{itemize}
    \item Diverse grading structures and course formats lead to variability in database schema, complicating schema linking.
    \item Lack of standardized datasets reflecting academic systems globally limits model generalization.
\end{itemize}

\subsection{Generalization Across Domains}
\begin{itemize}
    \item \textbf{Challenge}: Many text-to-SQL models struggle with domain adaptation, particularly when faced with unfamiliar database schemas or industries.
    \item \textbf{Future Direction}: Developing universal or multi-domain models that can interpret and generate accurate SQL for varied domains without additional retraining or extensive customization.
\end{itemize}

\subsection{Handling Ambiguity in Natural Language}
\begin{itemize}
    \item \textbf{Challenge}: Natural language queries can be ambiguous or lack specific details required for accurate SQL generation (e.g., missing parameters, vague terms).
    \item \textbf{Future Direction}: Implementing advanced disambiguation techniques or interactive clarification processes, where the model seeks clarification for ambiguous inputs to ensure query accuracy.
\end{itemize}

\subsection{Incorporating External Knowledge}
\begin{itemize}
    \item \textbf{Challenge}: Domain-specific queries often require contextual knowledge not present in the database schema, such as industry-specific terms or common data patterns.
    \item \textbf{Future Direction}: Integrating external knowledge bases or ontologies to enhance the model's contextual understanding and allow it to interpret queries with a broader scope.
\end{itemize}

\subsection{Optimizing SQL Efficiency}
\begin{itemize}
    \item \textbf{Challenge}: While accuracy is essential, the efficiency of SQL queries is also important, especially for large databases where complex queries can be resource-intensive.
    \item \textbf{Future Direction}: Researching optimization techniques for SQL generation, such as index-aware querying or using machine learning to predict and avoid performance bottlenecks in query structures.
\end{itemize}

\subsection{Human-in-the-Loop Systems}
\begin{itemize}
    \item \textbf{Challenge}: Fully automated systems may not meet the nuanced requirements of real-world applications.
    \item \textbf{Future Direction}: Developing interactive text-to-SQL systems where users can validate or edit SQL outputs, providing an extra layer of accuracy and flexibility for complex queries.
\end{itemize}

\subsection{Improved Interpretability and Debugging}
\begin{itemize}
    \item \textbf{Challenge}: Debugging and interpreting complex SQL statements generated by AI models can be challenging, particularly in mission-critical applications.
    \item \textbf{Future Direction}: Focusing on interpretability in model design, enabling end-users to understand how the generated SQL queries relate to the original natural language inputs, potentially through visual representations or explanations.
\end{itemize}

\begin{table}[!t]
\centering
\caption{Comparison of SQL and NoSQL Dataset Availability}
\renewcommand{\arraystretch}{1.2} 
\setlength{\tabcolsep}{8pt} 
\begin{tabular}{|p{3cm}|p{4cm}|}
\hline
\textbf{Category} & \textbf{Available Datasets} \\ \hline
Relational SQL & Spider, Spider 2.0, WikiSQL, CoSQL, BIRD \\ \hline
NoSQL & None (gap to be addressed) \\ \hline
Dialogue-Based SQL & CoSQL \\ \hline
Domain-Specific SQL & MIMICSQL (Healthcare), EDU-T5 (Education) \\ \hline
\end{tabular}
\label{tab:sql-nosql-comparison}
\end{table}

\subsection{Extending Text-to-SQL to NoSQL Databases}

NoSQL databases, such as MongoDB, Cassandra, and Redis, are crucial for handling unstructured, semi-structured, or rapidly changing data. Their flexible schema and distributed architecture make them essential for industries like finance, healthcare, e-commerce, and social networks, where scalability and real-time analytics are critical.

Despite advancements in Text-to-SQL systems for relational databases, there is a lack of models and datasets designed to query NoSQL systems, as shown in Table \ref{tab:sql-nosql-comparison}.

\begin{itemize}

\item \textbf{Challenge}: Existing benchmarks and models focus on SQL's structured schemas, while NoSQL databases require support for dynamic, document-based schemas. 

\item \textbf{Potential Solution}: Developing datasets tailored to NoSQL with JSON-like structures and models leveraging LLMs for unstructured data analytics \cite{dai2024}. This approach could bridge the gap between structured and unstructured data paradigms. \\

\end{itemize}

\section{Conclusion}
This paper concludes by highlighting the transformative potential of text-to-SQL systems in bridging the gap between natural language queries and database interactions, empowering non-technical users across various domains. While significant advancements have been achieved through AI-driven models and extensive benchmarks, critical challenges such as handling domain-specific complexities, extending support to NoSQL databases, and improving query efficiency remain. The survey underscores the need for future research to develop specialized datasets, enhance model generalization, and integrate contextual knowledge to address these gaps, thereby expanding the scope and utility of text-to-database technologies.

\bibliographystyle{IEEEtran}
\bibliography{ref.bib}

\begin{thebibliography}{10}
\providecommand{\url}[1]{#1}
\csname url@samestyle\endcsname
\providecommand{\newblock}{\relax}
\providecommand{\bibinfo}[2]{#2}
\providecommand{\BIBentrySTDinterwordspacing}{\spaceskip=0pt\relax}
\providecommand{\BIBentryALTinterwordstretchfactor}{4}
\providecommand{\BIBentryALTinterwordspacing}{\spaceskip=\fontdimen2\font plus
\BIBentryALTinterwordstretchfactor\fontdimen3\font minus \fontdimen4\font\relax}
\providecommand{\BIBforeignlanguage}[2]{{%
\expandafter\ifx\csname l@#1\endcsname\relax
\typeout{** WARNING: IEEEtran.bst: No hyphenation pattern has been}%
\typeout{** loaded for the language `#1'. Using the pattern for}%
\typeout{** the default language instead.}%
\else
\language=\csname l@#1\endcsname
\fi
#2}}
\providecommand{\BIBdecl}{\relax}
\BIBdecl

\bibitem{yu2018spider}
T.~Yu, R.~Zhang, K.~Yang \emph{et~al.}, ``Spider: A large-scale human-labeled dataset for complex and cross-domain semantic parsing and text-to-sql task,'' in \emph{Proceedings of the 2018 Conference on Empirical Methods in Natural Language Processing}, 2018, pp. 3911--3921.

\bibitem{shi2024survey}
L.~Shi, Z.~Tang, N.~Zhang \emph{et~al.}, ``A survey on employing large language models for text-to-sql tasks,'' \emph{arXiv preprint arXiv:2407.15186}, 2024.

\bibitem{singh2023}
A.~Singh, ``Exploring language models: A comprehensive survey and analysis,'' in \emph{2023 International Conference on Research Methodologies in Knowledge Management, Artificial Intelligence and Telecommunication Engineering (RMKMATE)}, 2023, pp. 1--4.

\bibitem{hong2024next}
Z.~Hong, Z.~Yuan, Q.~Zhang \emph{et~al.}, ``Next-generation database interfaces: A survey of llm-based text-to-sql,'' \emph{arXiv preprint arXiv:2406.08426}, 2024.

\bibitem{shi2024surveyemployinglargelanguage}
\BIBentryALTinterwordspacing
L.~Shi, Z.~Tang, N.~Zhang, X.~Zhang, and Z.~Yang, ``A survey on employing large language models for text-to-sql tasks,'' 2024. [Online]. Available: \url{https://arxiv.org/abs/2407.15186}
\BIBentrySTDinterwordspacing

\bibitem{lei2024spider}
\BIBentryALTinterwordspacing
F.~Lei, J.~Chen, Y.~Ye, R.~Cao, D.~Shin, H.~Su, Z.~Suo, H.~Gao, W.~Hu, P.~Yin, V.~Zhong, C.~Xiong, R.~Sun, Q.~Liu, S.~Wang, and T.~Yu, ``Spider 2.0: Evaluating language models on real-world enterprise text-to-sql workflows,'' 2024. [Online]. Available: \url{https://arxiv.org/abs/2411.07763}
\BIBentrySTDinterwordspacing

\bibitem{zhong2017seq2sql}
V.~Zhong, C.~Xiong, and R.~Socher, ``Seq2sql: Generating structured queries from natural language using reinforcement learning,'' \emph{arXiv preprint arXiv:1709.00103}, 2017.

\bibitem{li2023bird}
F.~Li, H.~Yu, X.~Li \emph{et~al.}, ``Can llm already serve as a database interface? a big bench for large-scale database grounded text-to-sqls,'' \emph{arXiv preprint arXiv:2305.03111}, 2023.

\bibitem{min2019pilotstudychinesesql}
\BIBentryALTinterwordspacing
Q.~Min, Y.~Shi, and Y.~Zhang, ``A pilot study for chinese sql semantic parsing,'' 2019. [Online]. Available: \url{https://arxiv.org/abs/1909.13293}
\BIBentrySTDinterwordspacing

\bibitem{lan2023unite}
W.~Lan, Z.~Wang, A.~Chauhan \emph{et~al.}, ``Unite: A unified benchmark for text-to-sql evaluation,'' \emph{arXiv preprint arXiv:2305.16265}, 2023.

\bibitem{yu2019CoSQL}
\BIBentryALTinterwordspacing
T.~Yu, R.~Zhang, H.~Y. Er, S.~Li, E.~Xue, B.~Pang, X.~V. Lin, Y.~C. Tan, T.~Shi, Z.~Li, Y.~Jiang, M.~Yasunaga, S.~Shim, T.~Chen, A.~R. Fabbri, Z.~Li, L.~Chen, Y.~Zhang, S.~Dixit, V.~Zhang, C.~Xiong, R.~Socher, W.~S. Lasecki, and D.~R. Radev, ``Cosql: A conversational text-to-sql challenge towards cross-domain natural language interfaces to databases,'' \emph{ArXiv}, vol. abs/1909.05378, 2019. [Online]. Available: \url{https://api.semanticscholar.org/CorpusID:202565697}
\BIBentrySTDinterwordspacing

\bibitem{xu2017sqlnet}
X.~Xu, C.~Liu, and D.~Song, ``Sqlnet: Generating structured queries from natural language without reinforcement learning,'' in \emph{International Conference on Learning Representations}, 2018.

\bibitem{yu2018typesql}
T.~Yu, Z.~Yao, Z.~Yang \emph{et~al.}, ``Typesql: Knowledge-based type-aware neural text-to-sql generation,'' in \emph{Proceedings of NAACL}, 2018, pp. 588--594.

\bibitem{GuoIRNet2019}
J.~Guo, Z.~Zhan, Y.~Gao, Y.~Xiao, J.-G. Lou, T.~Liu, and D.~Zhang, ``Towards complex text-to-sql in cross-domain database with intermediate representation,'' in \emph{Proceedings of the 57th Annual Meeting of the Association for Computational Linguistics (ACL)}.\hskip 1em plus 0.5em minus 0.4em\relax Association for Computational Linguistics, 2019.

\bibitem{raffel2020exploring}
C.~Raffel, N.~Shazeer, A.~Roberts \emph{et~al.}, ``Exploring the limits of transfer learning with a unified text-to-text transformer,'' \emph{Journal of Machine Learning Research}, vol.~21, pp. 1--67, 2020.

\bibitem{marshan2024medt5sql}
A.~Marshan, A.~N. Almutairi, A.~Ioannou, D.~Bell, A.~Monaghan, and M.~Arzoky, ``Medt5sql: a transformers-based large language model for text-to-sql conversion in the healthcare domain,'' \emph{Frontiers in Big Data}, vol.~7, p. 1371680, 2024.

\bibitem{hwang2019comprehensiveexplorationwikisqltableaware}
\BIBentryALTinterwordspacing
W.~Hwang, J.~Yim, S.~Park, and M.~Seo, ``A comprehensive exploration on wikisql with table-aware word contextualization,'' 2019. [Online]. Available: \url{https://arxiv.org/abs/1902.01069}
\BIBentrySTDinterwordspacing

\bibitem{wang2021ratsqlrelationawareschemaencoding}
\BIBentryALTinterwordspacing
B.~Wang, R.~Shin, X.~Liu, O.~Polozov, and M.~Richardson, ``Rat-sql: Relation-aware schema encoding and linking for text-to-sql parsers,'' 2021. [Online]. Available: \url{https://arxiv.org/abs/1911.04942}
\BIBentrySTDinterwordspacing

\bibitem{he2019xsqlreinforceschemarepresentation}
\BIBentryALTinterwordspacing
P.~He, Y.~Mao, K.~Chakrabarti, and W.~Chen, ``X-sql: reinforce schema representation with context,'' 2019. [Online]. Available: \url{https://arxiv.org/abs/1908.08113}
\BIBentrySTDinterwordspacing

\bibitem{lee2023ehrsql}
\BIBentryALTinterwordspacing
G.~Lee, H.~Hwang, S.~Bae, Y.~Kwon, W.~Shin, S.~Yang, M.~Seo, J.-Y. Kim, and E.~Choi, ``Ehrsql: A practical text-to-sql benchmark for electronic health records,'' \emph{arXiv preprint arXiv:2301.07695}, 2023. [Online]. Available: \url{https://arxiv.org/abs/2301.07695}
\BIBentrySTDinterwordspacing

\bibitem{qi2022rasatintegratingrelationalstructures}
\BIBentryALTinterwordspacing
J.~Qi, J.~Tang, Z.~He, X.~Wan, Y.~Cheng, C.~Zhou, X.~Wang, Q.~Zhang, and Z.~Lin, ``Rasat: Integrating relational structures into pretrained seq2seq model for text-to-sql,'' 2022. [Online]. Available: \url{https://arxiv.org/abs/2205.06983}
\BIBentrySTDinterwordspacing

\bibitem{scholak2021picardparsingincrementallyconstrained}
\BIBentryALTinterwordspacing
T.~Scholak, N.~Schucher, and D.~Bahdanau, ``Picard: Parsing incrementally for constrained auto-regressive decoding from language models,'' 2021. [Online]. Available: \url{https://arxiv.org/abs/2109.05093}
\BIBentrySTDinterwordspacing

\bibitem{ascoli2024esmmoderninsightsperspective}
\BIBentryALTinterwordspacing
B.~G. Ascoli, Y.~S.~R. Kandikonda, and J.~D. Choi, ``Esm+: Modern insights into perspective on text-to-sql evaluation in the age of large language models,'' 2024. [Online]. Available: \url{https://arxiv.org/abs/2407.07313}
\BIBentrySTDinterwordspacing

\bibitem{linkedin_sqlbot}
{LinkedIn Engineering Blog}, ``Practical text-to-sql for data analytics,'' \url{https://www.linkedin.com/blog/engineering/ai/practical-text-to-sql-for-data-analytics}, Oct. 2023, [Accessed: Dec. 29, 2024].

\bibitem{dai2024}
\BIBentryALTinterwordspacing
H.~Dai, B.~Y. Wang, X.~Wan, B.~Dai, S.~Yang, A.~Nova, P.~Yin, P.~M. Phothilimthana, C.~Sutton, and D.~Schuurmans, ``Uqe: A query engine for unstructured databases,'' 2024. [Online]. Available: \url{https://arxiv.org/abs/2407.09522}
\BIBentrySTDinterwordspacing

\end{thebibliography}
\end{document}